\documentclass[letterpaper,10pt,conference]{ieeeconf}

\IEEEoverridecommandlockouts
\usepackage{times}
\usepackage{float}
\usepackage{graphicx}
\usepackage{amsmath}
\usepackage{amssymb}
\usepackage{booktabs}
\usepackage[lined,boxed,ruled]{algorithm2e}
\usepackage[pagebackref=false,breaklinks=true,letterpaper=true,colorlinks,bookmarks=false]{hyperref}
\usepackage{cite}

\newcommand{\RR}[2]{\mathbb{R}^{#1 \times #2}}

\newcommand{\refEq}[1]{(\ref{#1})}
\newcommand{\refFig}[1]{Figure~\ref{#1}}

\newcommand{\refSec}[1]{Section~\ref{#1}}

\newcommand{\refTab}[1]{Table~\ref{#1}}

\def\bfx{{\boldsymbol{x}}}

\def\bfL{{{L}}}

\def\bfR{{{R}}}
\def\bfS{{{S}}}

\def\bfW{{{W}}}

\title{Human Motion Capture Using a Drone}
\author{Xiaowei Zhou, Sikang Liu, Georgios Pavlakos, Vijay Kumar, Kostas Daniilidis
\thanks{X. Zhou is with the the State Key Laboratory of CAD\&CG, Zhejiang University. Email address: xzhou@cad.zju.edu.cn.}
\thanks{S. Liu, G. Pavlakos, V. Kumar and K. Daniilidis are with the GRASP laboratory, University of Pennsylvania. Email addresses: \{sikang,pavlakos,kumar,kostas\}@seas.upenn.edu.}
}

\begin{document}

\maketitle
\thispagestyle{empty}
\pagestyle{empty}

\begin{abstract}
Current motion capture (MoCap) systems generally require markers and multiple calibrated cameras, which can be used only in constrained environments. In this work we introduce a drone-based system for 3D human MoCap. The system only needs an autonomously flying drone with an on-board RGB camera and is usable in various indoor and outdoor environments. A reconstruction algorithm is developed to recover full-body motion from the video recorded by the drone. We argue that, besides the capability of tracking a moving subject, a flying drone also provides fast varying viewpoints, which is beneficial for motion reconstruction. We evaluate the accuracy of the proposed system using our new DroCap dataset and also demonstrate its applicability for MoCap in the wild using a consumer drone.
\end{abstract}

\section{Introduction}


Capturing 3D human body motion is a challenging problem with many applications, e.g., in human-computer interaction, health care and sports. This problem has been conditionally solved by multi-camera motion capture (MoCap) systems (e.g. Vicon and Qualysis) in constrained studios. However, those MoCap systems suffer from their inflexibility and inconvenience: cameras require reliable fixation and frequent calibration, the tracking space is limited and fixed, and the subject should wear special markers. While being more challenging, image-based MoCap with an RGB camera has wider applicability and draws an increasing attention in recent years.

Despite the remarkable advances in monocular 3D human pose estimation (see the related-work section), these methods suffer from the inherent ambiguity of single-view reconstruction. The ambiguity is alleviated by learning a 3D pose prior from existing MoCap datasets but cannot be resolved geometrically. Another line of work aims to leverage multi-frame information in a video to reconstruct a nonrigid shape, which is known as nonrigid structure from motion (NRSFM) \cite{bregler2000recovering}. However, NRSFM requires sufficiently fast camera motion relative to the object \cite{akhter2011trajectory,park20113d}, which is impractical if the camera is fixed. 

\begin{figure}
	\includegraphics[width=\linewidth]{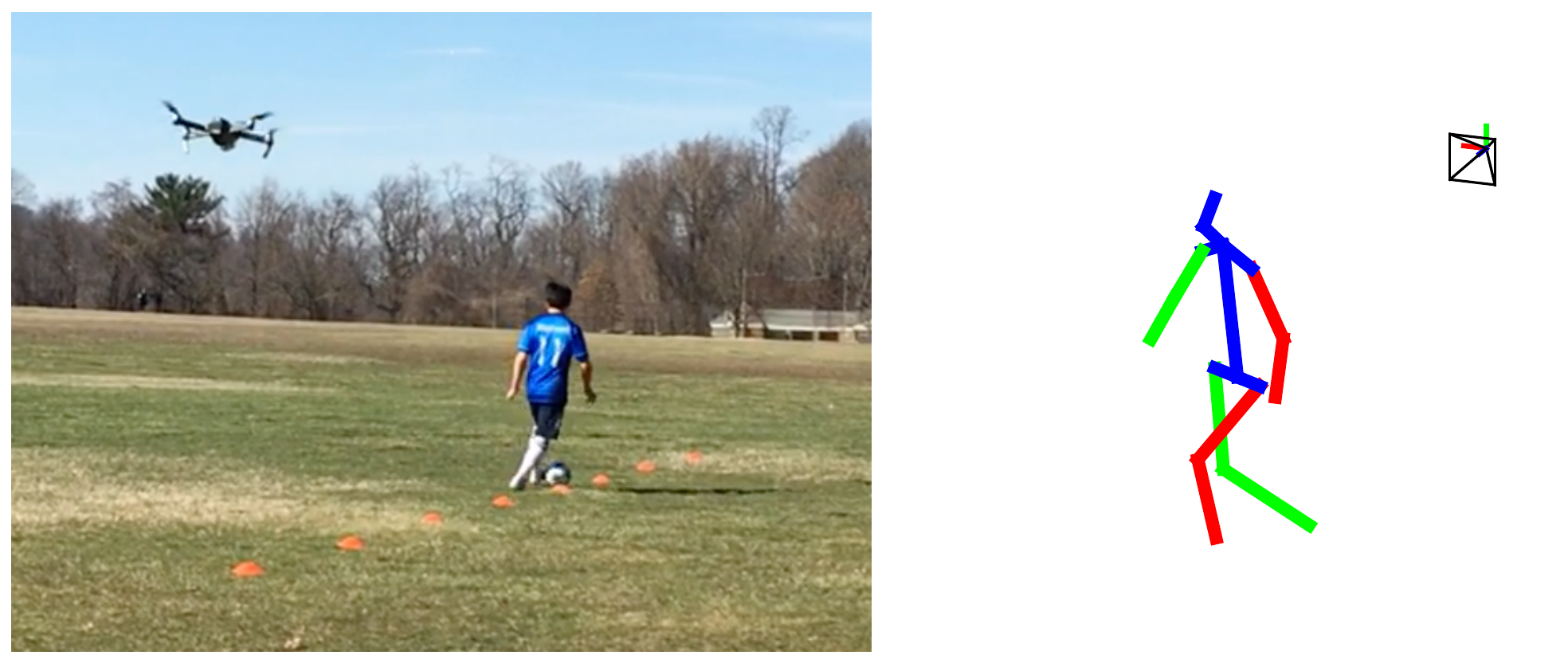}
   \caption{
   We propose a novel system for human motion capture based on an autonomously flying drone. (left) The drone orbits the human subject and records a video with an on-board RGB camera. The 3D full-body pose is recovered from the monocular video. (right) The reconstructed pose in the example frame is visualized at a novel viewpoint. The original viewpoint from the drone is indicated by the square pyramid. Please see the video at \url{https://github.com/daniilidis-group/drocap}.   
   }\label{fig:intro}
   \vspace{-1em}
\end{figure}

To address the above limitations of previous approaches, we propose a novel system for human body MoCap using a drone (see \refFig{fig:intro})
leveraging the state-of-the-art techniques in autonomous drones and computer vision. 
An autonomously flying drone orbits and records a video of the subject, providing fast varying viewpoints about the subject.
A convolutional neural network (CNN) based 2D pose estimator produces reliable 2D tracks of body joints from the video, 
which are input to a 3D pose estimator that robustly initializes reconstruction and suppresses outliers in the 2D tracks.
Finally, a NRSFM algorithm is developed to further refine the reconstruction using sequence information and impose the articulation constraint of human body.  

Our contributions are summarized below:

\begin{itemize}
\item We propose a novel drone-based system for human MoCap, which is simple (uses only a drone with an on-board RGB camera), flexible (works both indoors and outdoors) and readily usable (needs no particular system calibration or model training). 
\item We argue that, compared to using a static camera, using a drone for video recording is able to provide fast camera motion relative to the subject, which is necessary for motion reconstruction.   
\item We develop a reconstruction algorithm to recover 3D human poses from the drone-based video, which consists of a novel synthesis between single-view pose estimation and sequence-based NRSFM for both robustness and accuracy.  
\item We introduce a novel rank-1 constraint to model the articulation of human body. The proposed constraint is effective and applicable even if the limb lengths of the subject are unknown.
\item We release a drone-based MoCap dataset named DroCap, which consists of human motion videos captured by a flying drone and ground truth obtained by a Vicon MoCap System. The dataset is available at \url{https://github.com/daniilidis-group/drocap}.
\end{itemize}

\subsection*{Related work}\label{sec:related}

Markerless human motion capture has been a long standing problem in computer vision. Early work in this area was focused on model-based body tracking in monocular \cite{sminchisescu2003kinematic} or multi-view sequences \cite{gall2010optimization}, which in general requires initialization in the first frame and is prone to tracking failures. To address these issues, more robust bottom-up approaches were proposed. These approaches first detect 2D body joints in images and lift them to 3D by assuming a given skeleton \cite{taylor2000reconstruction,mori2006recovering,park20113d}, searching MoCap databases \cite{yasin2016dual} or learning statistical models from MoCap data, such as the loosely-limbed model \cite{sigal2012loose}, principal component analysis \cite{simo2012single,simo2013joint,zhou2014spatio}, sparse representation \cite{ramakrishna2012reconstructing,akhter2015pose,zhou2015sparse} and body shape models \cite{guan2009estimating,bogo2016keep}. The main limitation of these approaches is that single-view reconstruction is inherently ill-posed and using geometric priors alone is insufficient to resolve the reconstruction ambiguity. Moreover, the pose prior learned from training data might not be able to explain the pose variability in testing examples resulting in inaccurate reconstructions. While the reconstruction ambiguity can be resolved by using multiple cameras \cite{belagiannis20143d,puwein2014joint,elhayek2015efficient}, the calibrated multi-camera system is not easily accessible in practice.

Another strand of work directly learns the mapping from an image to 3D pose parameters using annotated training data in the form of image-pose pairs, which is referred to as discriminative methods. The advantage of discriminative methods is their ability to leverage image cues, such as shading and occlusion, to reduce the reconstruction ambiguity. A variety of supervised learning methods have been adopted ranging from traditional techniques such as linear regression \cite{agarwal2006recovering}, kernel density estimation \cite{ionescu2014human} and Gaussian processes \cite{bo2010twin} to modern deep convolutional neural networks (CNNs) \cite{li2015maximum,tekin2015predicting}. The main limitation for the discriminative methods is that they require a large number of training images with corresponding 3D pose annotations, which could only be collected using MoCap systems. While there have been large-scale MoCap datasets, such as HumanEva \cite{sigal2010humaneva} and Human3.6M \cite{ionescu2014human}, they lack diversity in scenes, subject appearances and actions. As a consequence, the models trained on indoor MoCap datasets are prone to overfitting and can hardly be applied to outdoor images. Some recent works tried to address the scarcity of training data by image synthesis \cite{chen2016synthesizing,rogez2016mocap}, but the performance of a model trained on synthesized images is not guaranteed when applied to real images due the difference in image statistics. 

Human motion capture is closely related to nonrigid structure from motion (NRSFM) \cite{bregler2000recovering,paladini2009factorization,dai2012simple,garg2013dense,wandt20163d}. In NRSFM, a deformable shape is reconstructed from 2D keypoint tracks extracted from a video.
Most of the state-of-the-art NRSFM methods assume a low-rank prior on the deformable shape over time. But unlike the single-view methods that learn bases from MoCap data (e.g. \cite{simo2012single,ramakrishna2012reconstructing}), NRSFM methods recover the bases during reconstruction without the need of training data, which might better fit the subject.   
However, it is difficult to apply existing NRSFM pipelines for human motion capture. 
First, obtaining clean 2D keypoint tracks from a video is difficult especially for fast moving and deformable objects like human body. 
Second, NRSFM requires fast camera motion \cite{akhter2011trajectory,park20113d}, which is impractical in usual scenarios. 
In this work, we leverage NRSFM but combine it with single-view pose estimation for robust initialization and outlier handling. 
Moreover, using an on-board camera on a drone for video recording naturally provides fast camera motion. 

Using flying cameras for human MoCap was proposed in \cite{xu2016flycap}, in which the system consists of multiple drones equipped with RGB-D sensors and solves the problem by model-based tracking.  However, it requires a scanning stage in which the subject stays static for body shape scanning using depth sensors. Also, RGB-D sensors are restricted to indoor environments and the sensing range is limited. 
Compared to \cite{xu2016flycap}, the proposed system is more widely usable. It only needs a consumer drone with an RGB camera, doesn't require any initialization or scanning stage, and works in both indoor and outdoor environments.

\section{Approaches}

An autonomously flying drone is used for data collection. The drone tracks and orbits the subject with an on-board RGB camera pointing at the subject and recording a video. This functionality has been implemented in many commercial drones such as DJI Phantom 4 and Mavic Pro. The motivation for using a drone instead of a fixed camera for video recording is the capability to provide a sequence of fast varying viewpoints about the subject, which is favorable to motion reconstruction. The importance of camera motion will be experimentally demonstrated in \refSec{sec:rotation}. 

Given a monocular video of the subject recorded from the orbiting drone, we aim to recover the 3D pose sequence of the subject. Our pipeline consists of the following steps: 1) \emph{2D pose detection} in which the subject is detected and the 2D pose is estimated in each frame; 2) \emph{single-frame initialization} in which the camera viewpoints and 3D human poses are initialized by the single-view pose estimation method \cite{zhou2016sparseness}; 3) \emph{multi-frame bundle adjustment} in which the camera viewpoints and 3D poses are refined by minimizing the nuclear norm of shape matrix with an articulation constraint. This pipeline is analogous to the successful experience in rigid structure from motion: we detect keypoints, incrementally reconstruct each frame and optimize all unknowns in bundle adjustment. 

\subsection{2D pose detection}

The bounding box of the subject in each frame is obtained by either object tracking or detection. For example, the DJI Mavic Pro comes with the active tracking feature and provides the bounding box of the tracked subject. Otherwise, an object detector can be applied to localize the subject in each frame, e.g., the faster R-CNN \cite{ren2016faster} in our experiments.  

We adopt the stacked hourglass model proposed by Newell et al. \cite{newell2016stacked} for 2D pose estimation, which is the state-of-the-art method for this problem.
It is a fully convolutional neural network (F-CNN) with two stacked hourglass components, each of which consists of a series of downsampling and upsampling layers implementing the bottom-up and top-down processing to integrate local image feature with global context over the whole image. The input to the network is a 2D image with a bounding box around the subject and the output is a set of heat maps with each showing the predicted spatial distribution of the corresponding joint. The details are referred in \cite{newell2016stacked}. 

\subsection{Single-frame initialization}

After the 2D body joints are located in the image sequence, NRSFM methods can be used to reconstruct the 3D poses from the 2D tracks of body joints. However, there are two difficulties for this approach. First, there might be a considerable number of gross errors (outliers) in the detected 2D joints due to occlusion, left-right ambiguity and background clutters. The existing NRSFM methods can hardly handle outliers as NRSFM is an ill-posed problem without much information to correct the input error. Secondly, NRSFM methods typically rely on low-rank factorization which requires a predefined rank while the best rank is often unknown. 

To address these difficulties, we propose to use a recent method for single-view 3D pose estimation \cite{zhou2016sparseness} to initialize the reconstruction. In \cite{zhou2016sparseness}, a pose dictionary is learned from existing MoCap data and the pose to be reconstructed is assumed as a linear combination of the bases in the dictionary. In this way, the number of unknowns is reduced and the optimization is easier to solve compared to NRSFM where the bases are also unknown. An EM algorithm is also developed in \cite{zhou2016sparseness} to account for uncertainties in CNN based 2D joint localization by incorporating the 3D pose prior learned from MoCap data. Even if the learned pose dictionary cannot perfectly represent the poses to be reconstructed, this  step can reliably initialize the camera viewpoints and correct outliers in the 2D input.  

\subsection{Multi-frame bundle adjustment}

A downside of the single-view reconstruction is that the pose bases learned from other MoCap datasets might not be able to represent the new poses in test images. We propose to solve this problem by adapting the pose bases to the current sequence, which has been realized in NRSFM where a low-rank basis is learned along with other unknowns from 2D keypoint tracks. 

We adopt the nuclear norm minimization scheme which has been used in many recent NRSFM methods (e.g. \cite{dai2012simple,garg2013dense}). Compared to factorization based methods, the advantages are two-fold: 1) there is no need to explicitly define a rank; and 2) nuclear norm minimization is convex. We additionally propose a novel rank-1 constraint to model the articulated nature of human body. 

\subsubsection{Objective function}

Suppose the 2D pose and 3D pose of the subject in frame $t$ are represented by $\bfW_t\in\RR{2}{p}$ and $\bfS_t\in\RR{3}{p}$ respectively, where $p$ is the number of joints. Following the general practice in NRSFM (e.g. \cite{bregler2000recovering,dai2012simple}), we assume that an orthographic camera model is used and both $\bfW_t$ and $\bfS_t$ are centralized, such that
\begin{align}
\bfW_t = \bfR_t\bfS_t
\end{align}
where $\bfR_t\in\RR{2}{3}$ denotes the first two rows of the camera rotation matrix at frame $t$. 

Given the 2D pose sequence $\bfW=\{\bfW_1,\cdots,\bfW_n\}$, we recover the 3D pose sequence $\bfS=\{\bfS_1,\cdots,\bfS_n\}$ and the camera rotation sequence $\bfR=\{\bfR_1,\cdots,\bfR_n\}$ by solving the following optimization problem:
\begin{align}\label{eq:cost}
\min_{\bfS,\bfR,\bfL} f(\bfS,\bfR,\bfL) + \alpha \|\bfS^{\#}\|_*
\end{align}
where $f(\bfS,\bfR,\bfL)$ is a smooth function composed of the following terms:

\begin{align}\label{eq:smooth}
f(\bfS,\bfR,\bfL) = \sum_{t=1}^n \| \bfW_t - \bfR_t\bfS_t \|_F^2 + \gamma\|\ell(\bfS) - \bfL\|_F^2
\end{align}

The first term in \refEq{eq:smooth} is the sum of reprojection errors over all joints in all frames. The second term enforces the articulation (anthropomorphic) constraint, i.e., the limb lengths should be constant across the sequence. However, as the scale of the reconstructed 3D structure is determined by the given 2D structure under the orthographic projection, the size of the reconstructed structure may vary in different frames depending on the distance from the camera to the subject. Therefore, instead of constraining limb lengths as constants, we enforce that the the ratios between limb lengths to be unchanged across frames. Suppose $\ell(\cdot)$ is a function such that the $t$-th column of $\ell(\bfS)$ gives the squared limb lengths of $\bfS_t$, $\ell(\bfS)$ should be rank-1 if the articulation constraint is satisfied. To simplify the optimization, we minimize the difference between $\ell(\bfS)$ and an auxiliary rank-1 matrix $\bfL$ instead of directly constraining the rank of $\ell(\bfS)$. $\bfL$ is also unknown and updated during optimization. Note that $\ell(\bfS)$ gives the squared lengths which are differentiable.

The second term in \refEq{eq:cost} is a nonsmooth regularizer that enforces the low-rankness of the reconstructed poses, where $\|\cdot\|_*$ is the nuclear norm and $\bfS^{\#}$ denotes a rearrangement of $\bfS$ such that the $t$-th column of $\bfS^{\#}$ is the vectorized $\bfS_t$. When $\gamma=0$, the formulation \refEq{eq:cost} is identical to the ones used in previous work for NRSFM (e.g. \cite{dai2012simple,garg2013dense}).

Note that the temporal smoothness of both $\bfS_t$ and $\bfR_t$ could be conveniently imposed by minimizing their temporal changes. We didn't include them in \refEq{eq:cost} for simplicity and observed that adding them didn't significantly improve the quantitative results.

\subsubsection{Optimization}

The problem in \refEq{eq:cost} is nonlinear and nonconvex. However, the single-frame initialization stage has provided reliable initialization, which allows us to solve the problem in \refEq{eq:cost} by local optimization. 

More specifically, we alternately update each variable while fixing the others. The camera rotation $\bfR$ can be updated with any parameterization of rotation matrix. In accordance with the initialization method \cite{zhou2016sparseness}, we optimize $\bfR$ over the Stiefel manifold, which is implemented using the manifold optimization toolbox \cite{boumal2014manopt}. The update of $\bfL$ is a standard low-rank approximation problem which can be analytically solved by singular value decomposition. 
The update of $\bfS$ is more complicated where the objective consists of a smooth loss function and a nonsmooth nuclear norm regularizer. We adopt the proximal gradient (PG) method \cite{nesterov2007gradient}, which updates $\bfS$ iteratively according to the following rule until convergence:
\begin{align}\label{eq:pg}
    \bfS^{k+1}=\arg\min_{\bfS}~\frac{1}{2}\left\|\bfS-[\bfS^k-\frac{1}{\mu}\nabla f_{\bfS^{k}}]\right\|_F^2 + \frac{\alpha}{\mu}\left\|\bfS^{\#}\right\|_*
\end{align}
where $\nabla f_{\bfS^{k}}$ is the gradient of the smooth function $f$ evaluated at the previous estimate $\bfS^{k}$ and $\mu$ determines the step size. The subproblem in \refEq{eq:pg} can be analytically solved by the singular value thresholding \cite{cai2010singular}. To additionally accelerate the convergence of the PG iterations, the Nesterov acceleration scheme \cite{nesterov2007gradient} is also implemented.

\section{Results}

\subsection{Importance of camera motion}\label{sec:rotation}

We first demonstrate that fast camera motion is favorable to motion reconstruction, which is the motivation of using drone for data collection in the proposed system. To achieve an arbitrary camera velocity, we use synthetic 2D input by projecting groundtruth 3D body joints to 2D with a virtual orthographic camera rotating around the subject. The 3D data is from Human3.6M \cite{ionescu2014human}, a large-scale MoCap dataset. All sequences of 15 actions from Subject 9 and 11 are used for evaluation. The sequences are subsampled at a frame rate of 24 fps and the first 10 seconds of each sequence are used for evaluation. The reconstruction error is used as the error metric:
\begin{align*}
e = \min_{\mathcal{T}}\frac{1}{p}\sum_{i=1}^p\|\hat\bfx_i-\mathcal{T}(\bfx_i^*)\|_2,
\end{align*}
which calculates the mean distance between the estimated joint locations $\hat\bfx$ and ground truth $\bfx^*$ after a similarity transformation $\mathcal{T}$ to align them.

\begin{figure}
  \centering
  \includegraphics[width=0.6\linewidth]{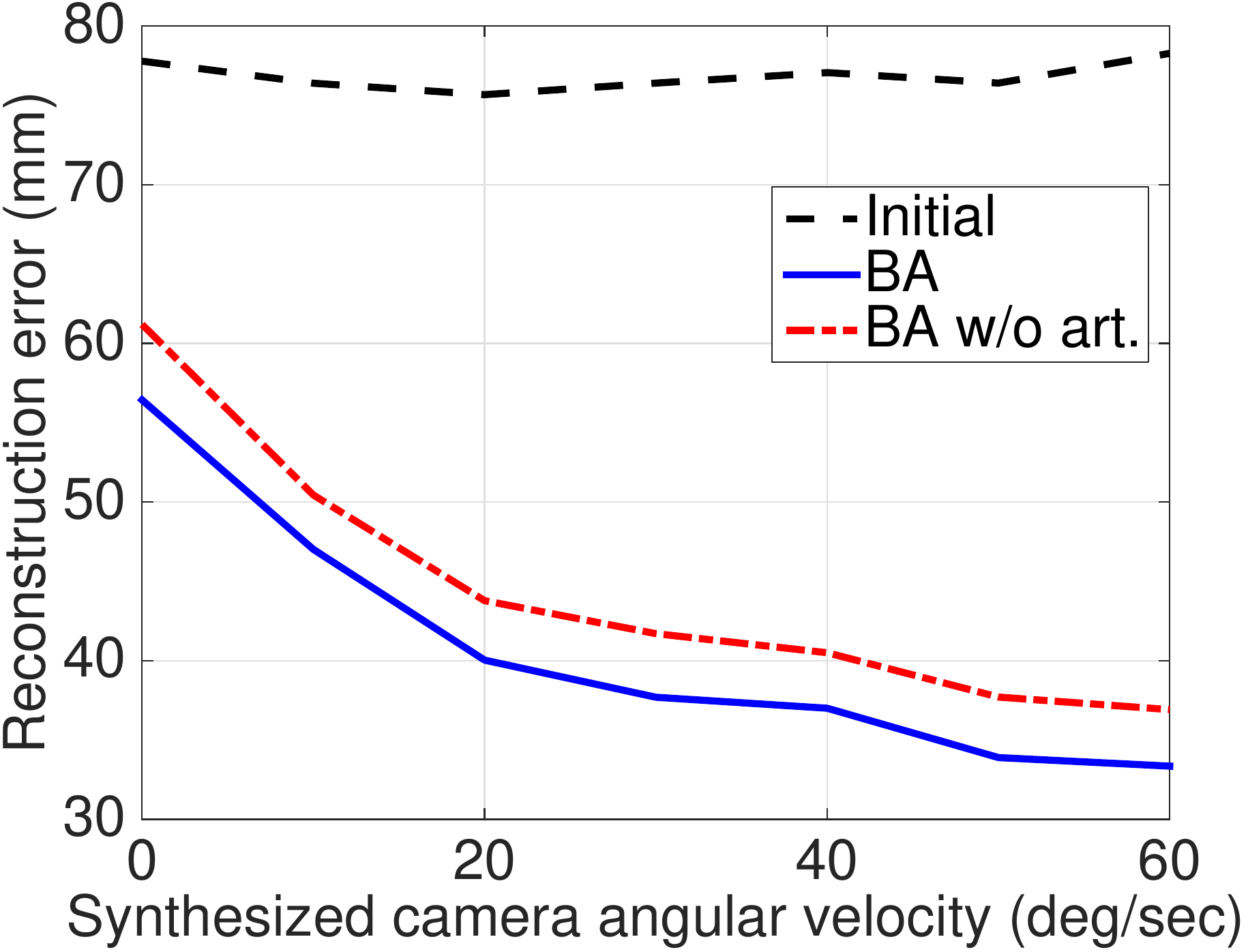}\\
  \vspace{1em}
  \caption{The 3D reconstruction error as a function of the angular velocity of virtual camera. The three curves correspond to the single-view initialization (Initial) by \cite{zhou2016sparseness}, the multi-frame bundle adjustment (BA) and BA without the articulation constraint (BA w/o art.). }\label{fig:rotation}
\end{figure}

The mean reconstruction error averaged over all sequences is plotted in \refFig{fig:rotation} as a function of the angular velocity of the virtual camera. The results of the initial single-view method \cite{zhou2016sparseness} and the bundle adjustment with ($\gamma=1$) and without ($\gamma=0$) the articulation constraint are presented. 
To avoid training on the same dataset, we learn a pose dictionary from the CMU Mocap dataset \cite{mocap} for single-frame initialization, achieving a mean error around $77$mm. If the pose dictionary is learned from the same dataset (the training set of Human3.6M), the mean error is around $48$mm, but this setting is impractical in real applications.  
\refFig{fig:rotation} shows that the multi-frame bundle adjustment improves upon the initial single-view estimation and the accuracy becomes better as the camera rotates faster, which validates that the multi-view information from fast varying viewpoints helps motion reconstruction.  
The benefit of imposing the articulation constraint is also clearly demonstrated.

\begin{figure*}
  \centering
  \includegraphics[width=\linewidth]{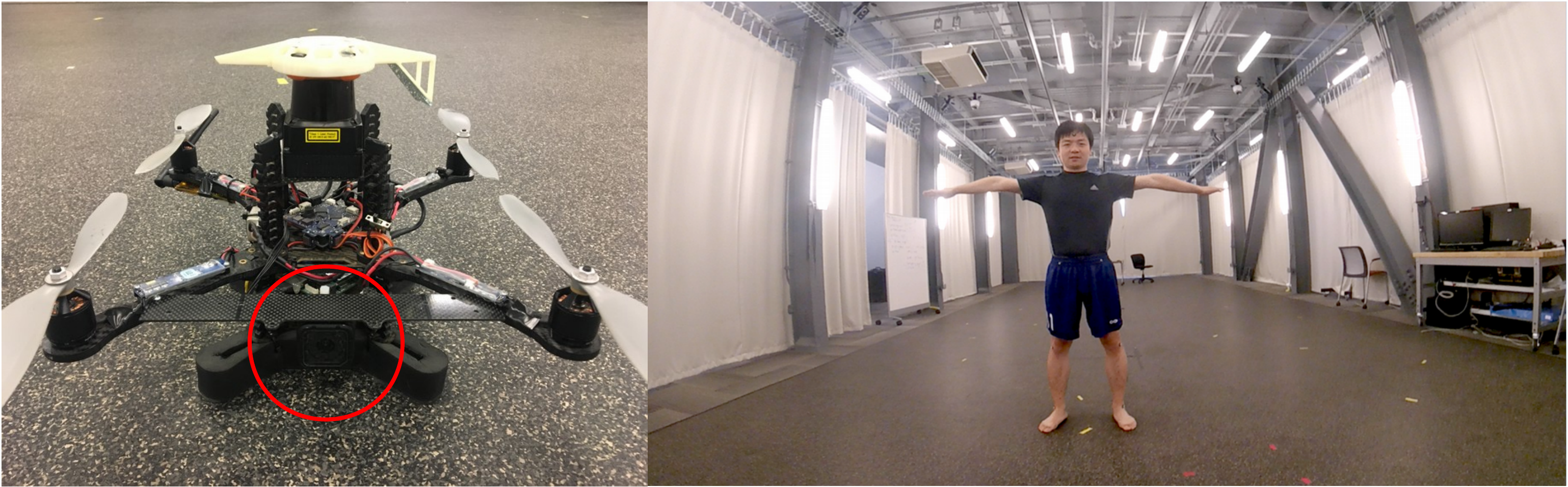}
  \caption{Left: the AscTec Pelican quadrotor platform used for data collection (the red circle labels the on-board GoPro camera). Right: a sample video frame from the on-board camera. }\label{fig:lab}
\end{figure*}
\vspace{1em}
\begin{figure*}
  \centering
  \includegraphics[width=\linewidth]{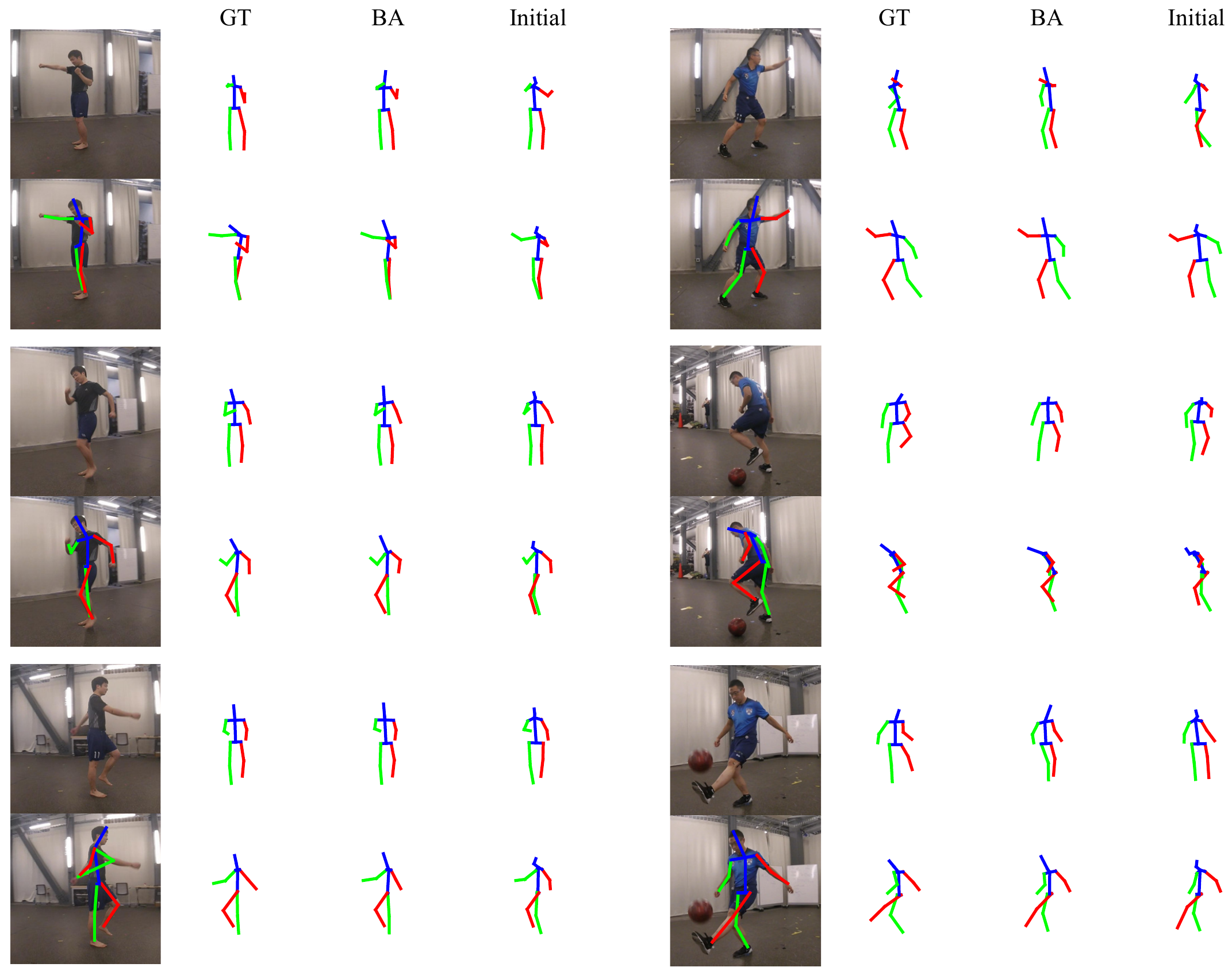}
  \caption{Qualitative evaluation on the DroCap dataset. Each panel corresponds to an example frame. For each panel, the 1st column shows the cropped image (top) and estimated 2D pose (bottom). The 2nd to the 4th columns correspond to the groundtruth poses (GT), reconstructions after bundle adjustment (BA) and from the initialization (Initial), respectively, visualized in front and profile views.}\label{fig:DroCap}
\end{figure*}

\begin{figure*}
\begin{minipage}{0.33\textwidth}
  \includegraphics[width=.99\linewidth]{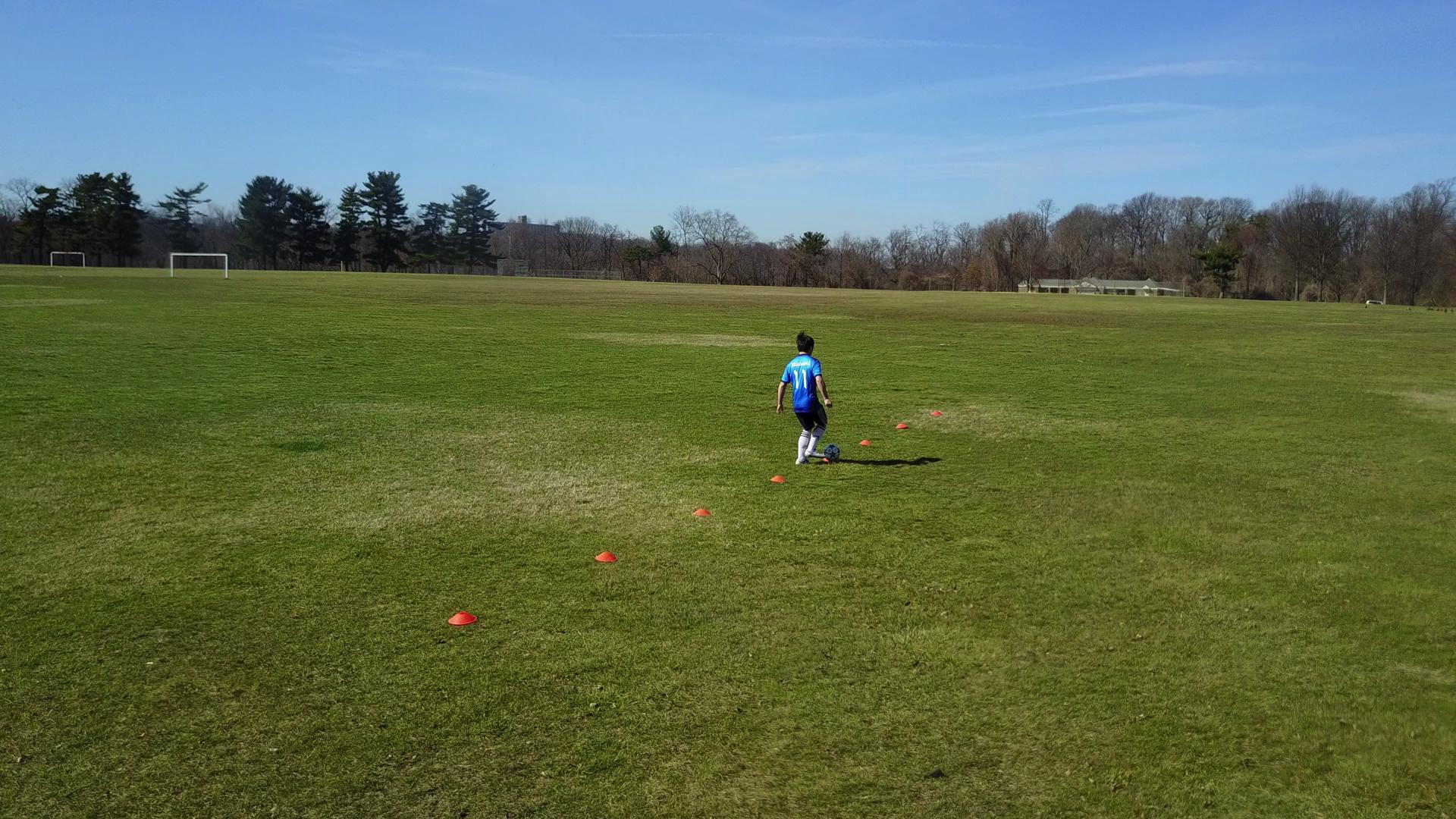}\\
  \includegraphics[width=.99\linewidth]{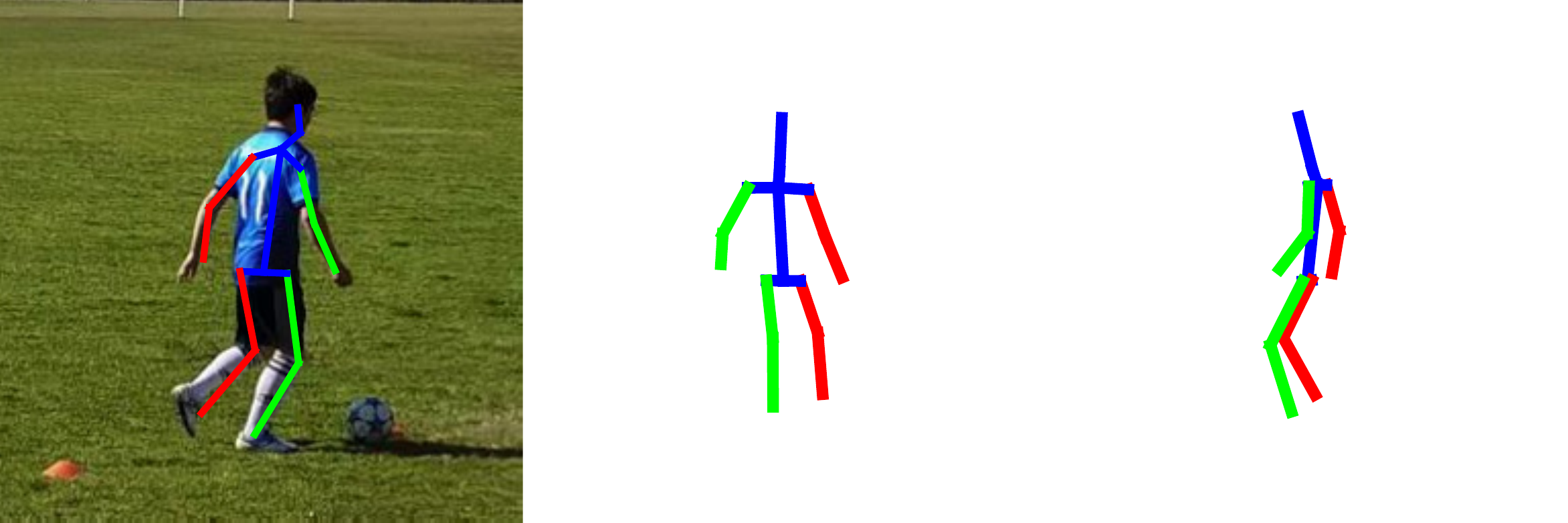}\\
  \includegraphics[width=.99\linewidth]{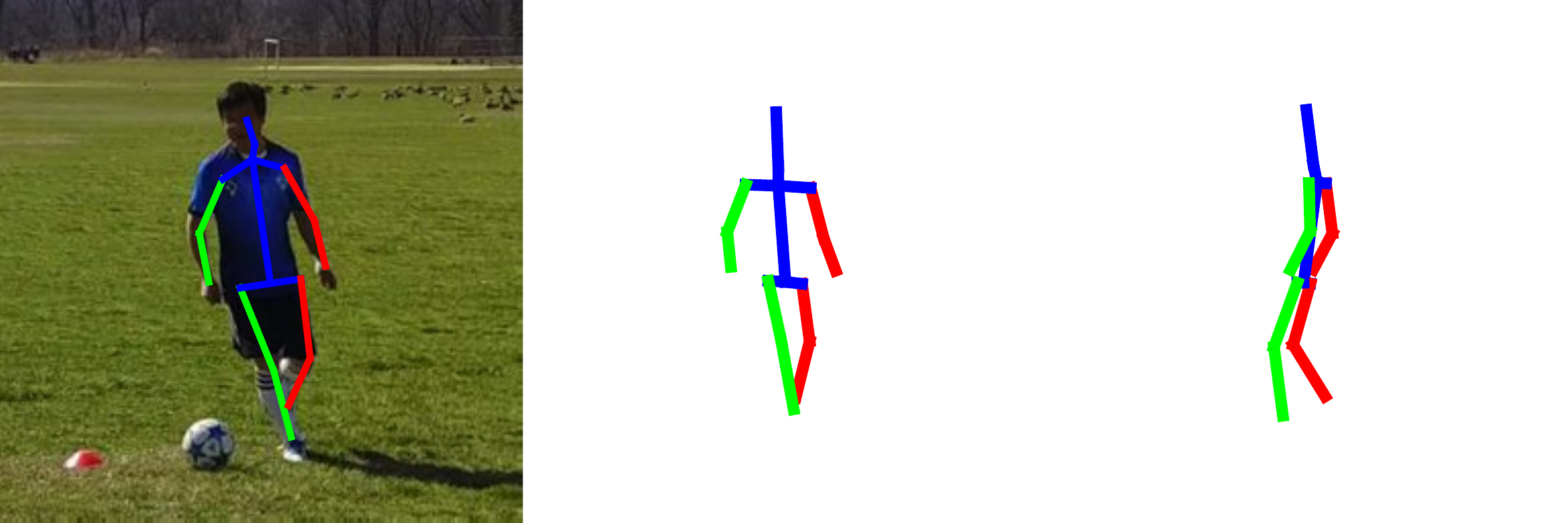}\\
  \includegraphics[width=.99\linewidth]{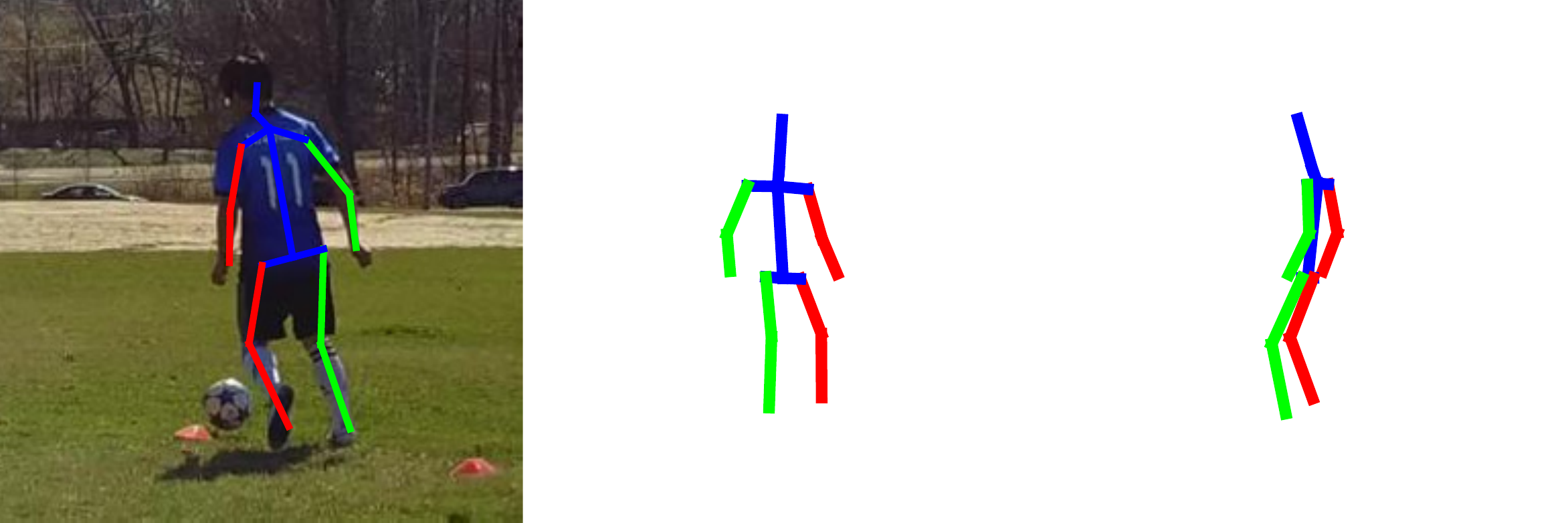}\\
  \includegraphics[width=.99\linewidth]{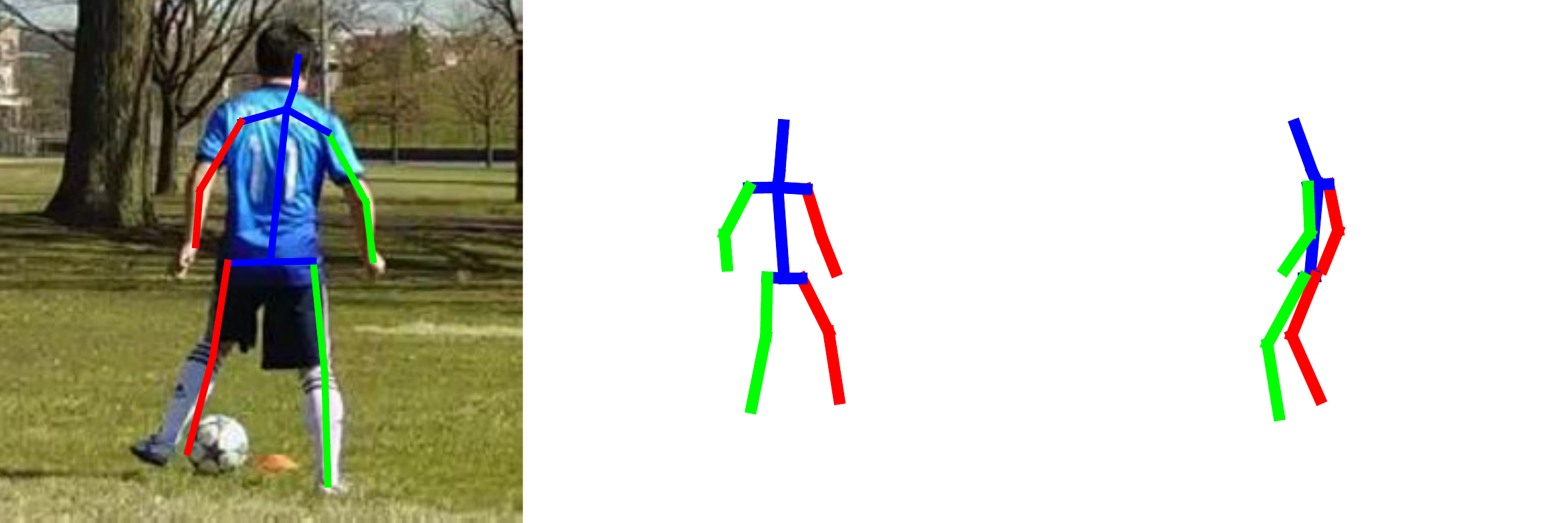}\\
  \includegraphics[width=.99\linewidth]{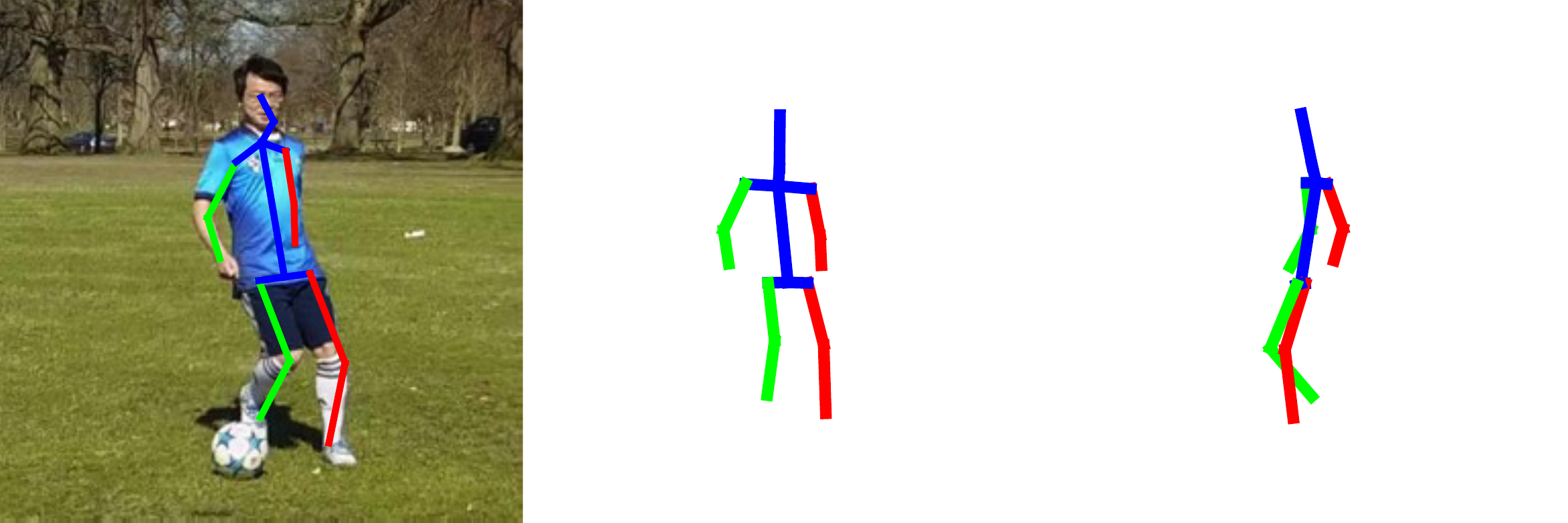}\\
\end{minipage}
\begin{minipage}{0.33\textwidth}
  \includegraphics[width=.99\linewidth]{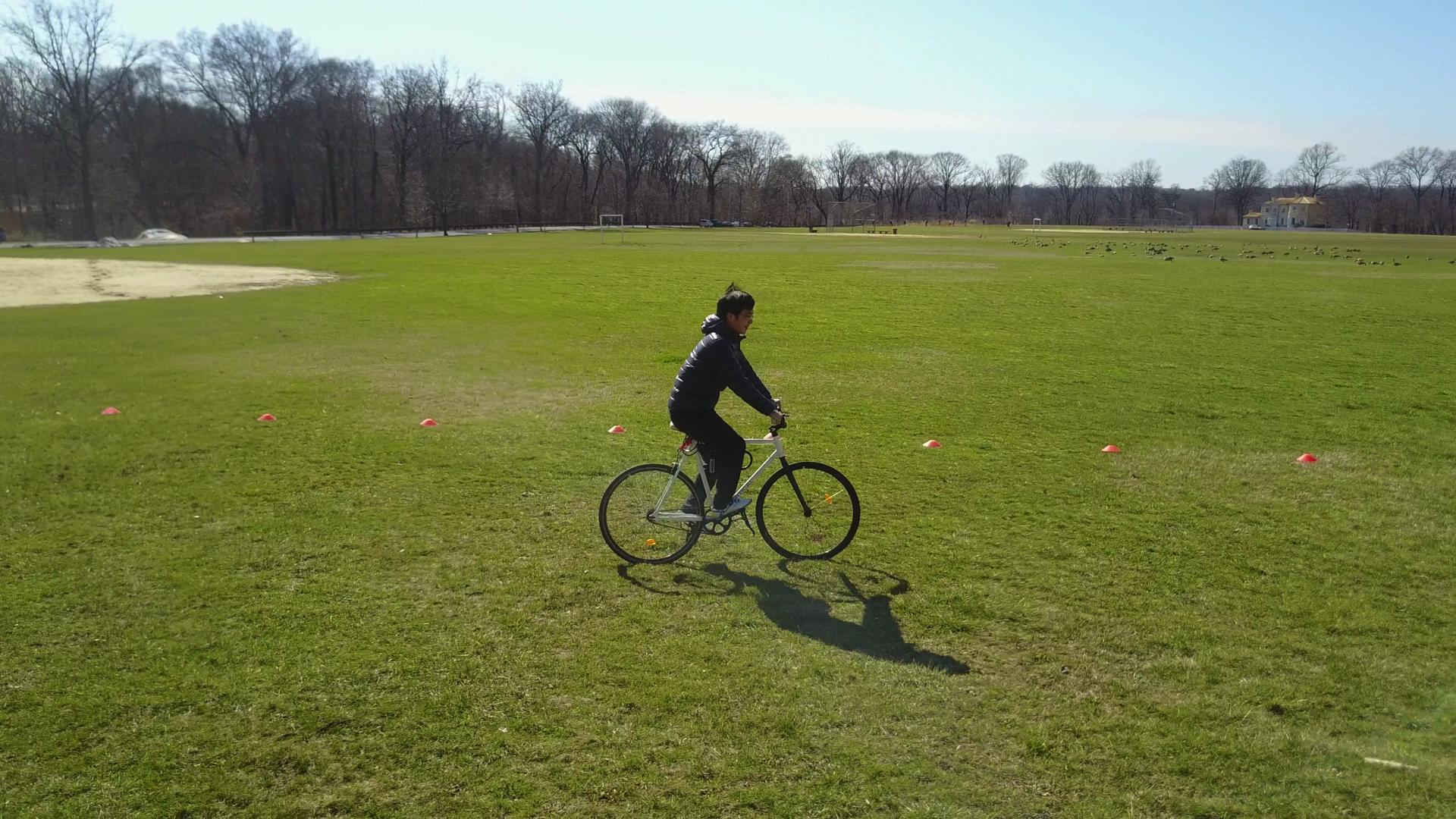}\\
  \includegraphics[width=.99\linewidth]{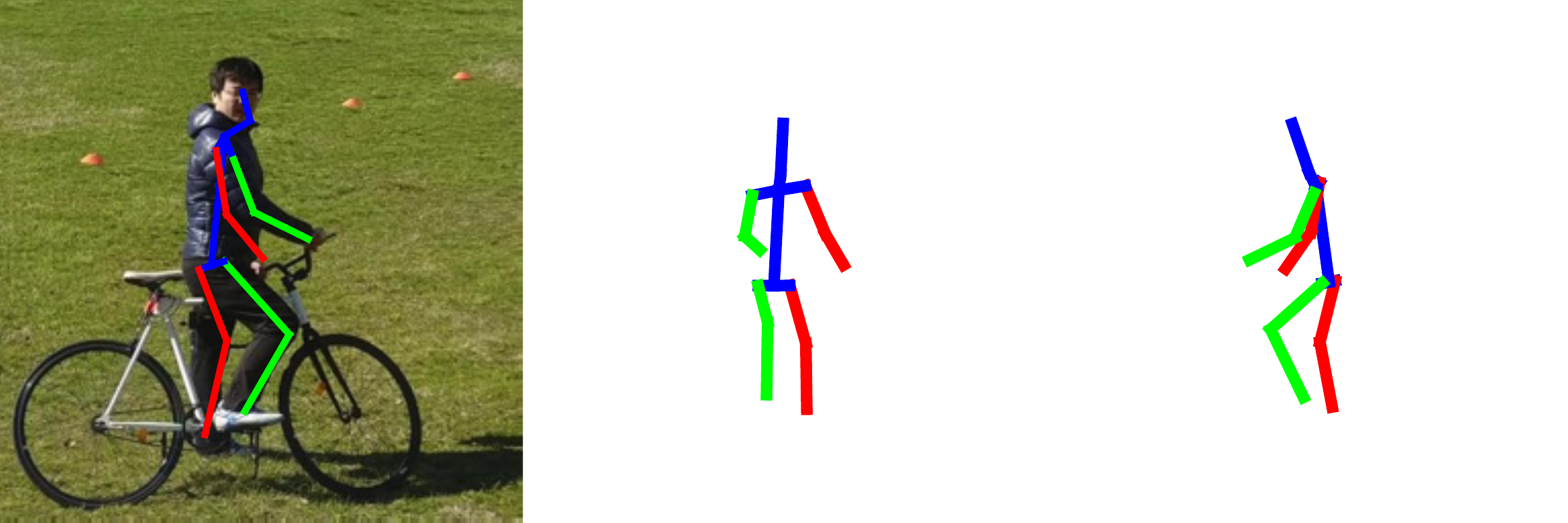}\\
  \includegraphics[width=.99\linewidth]{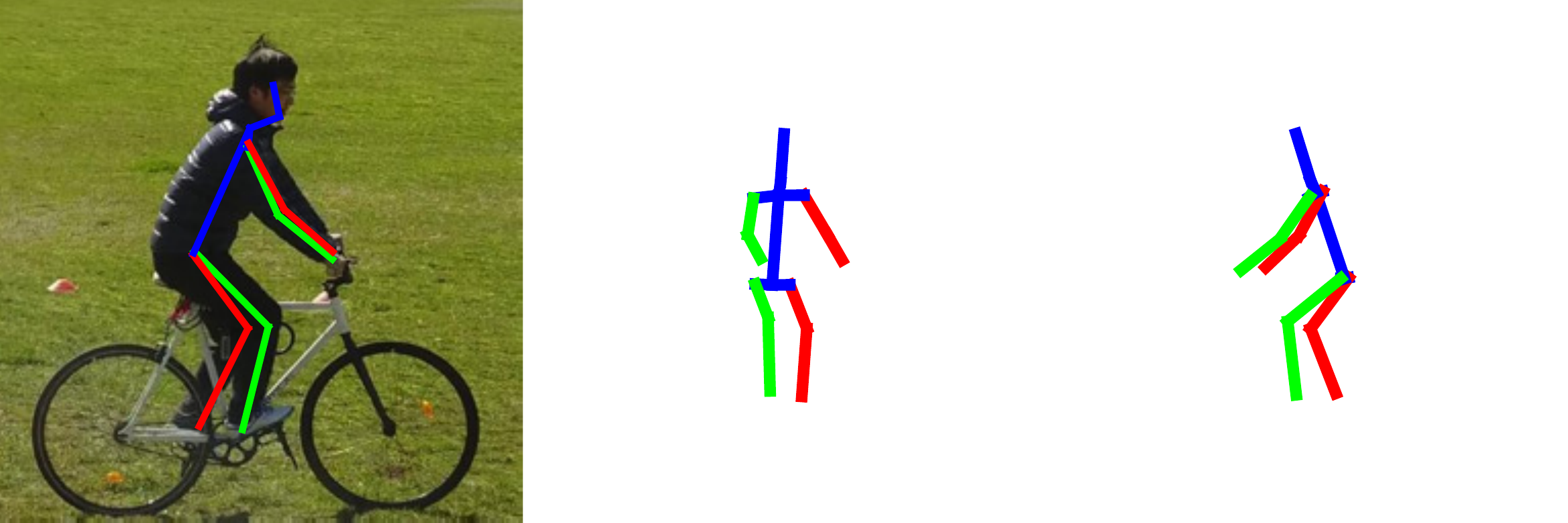}\\
  \includegraphics[width=.99\linewidth]{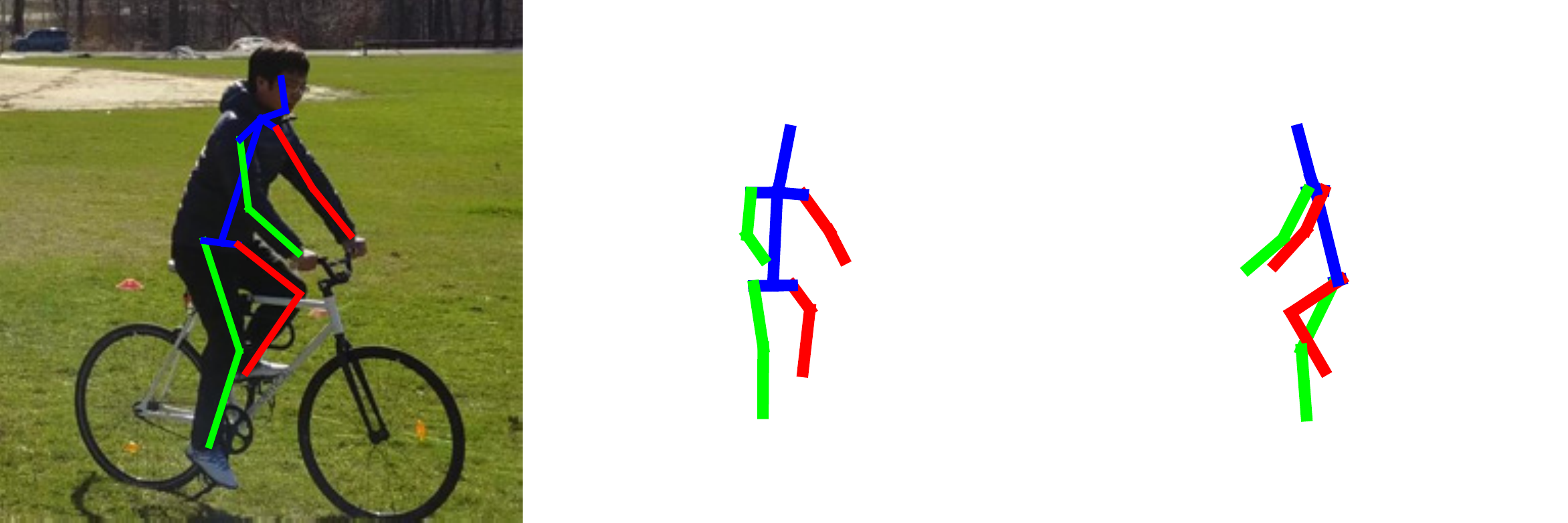}\\
  \includegraphics[width=.99\linewidth]{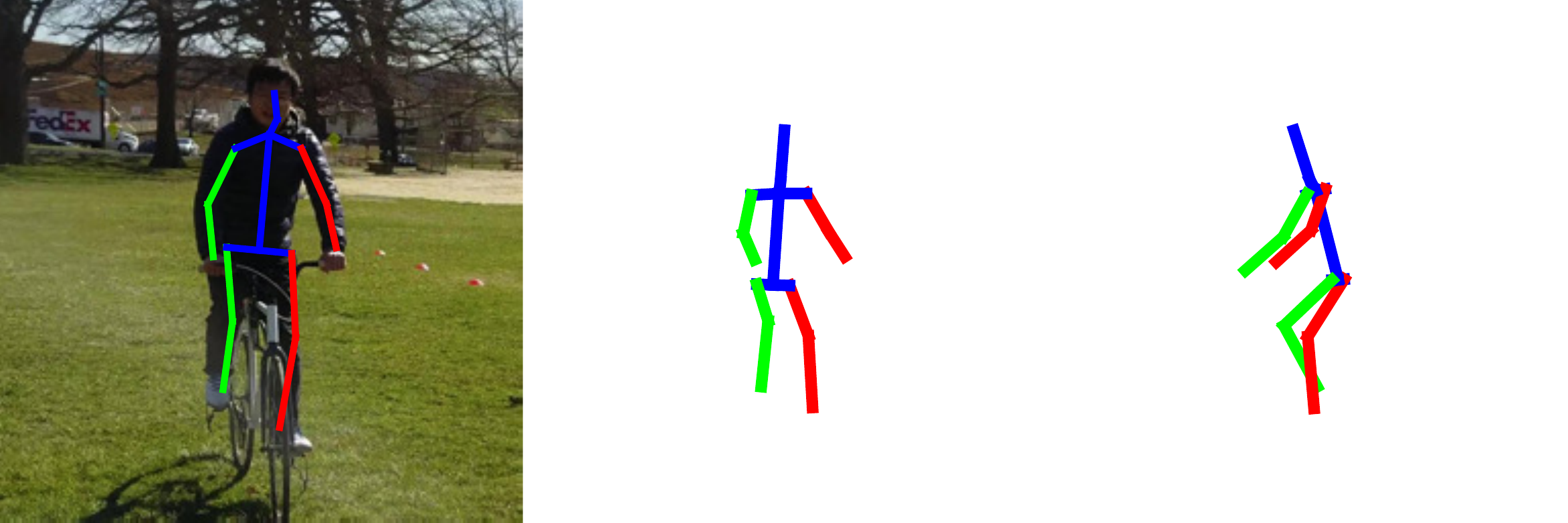}\\
  \includegraphics[width=.99\linewidth]{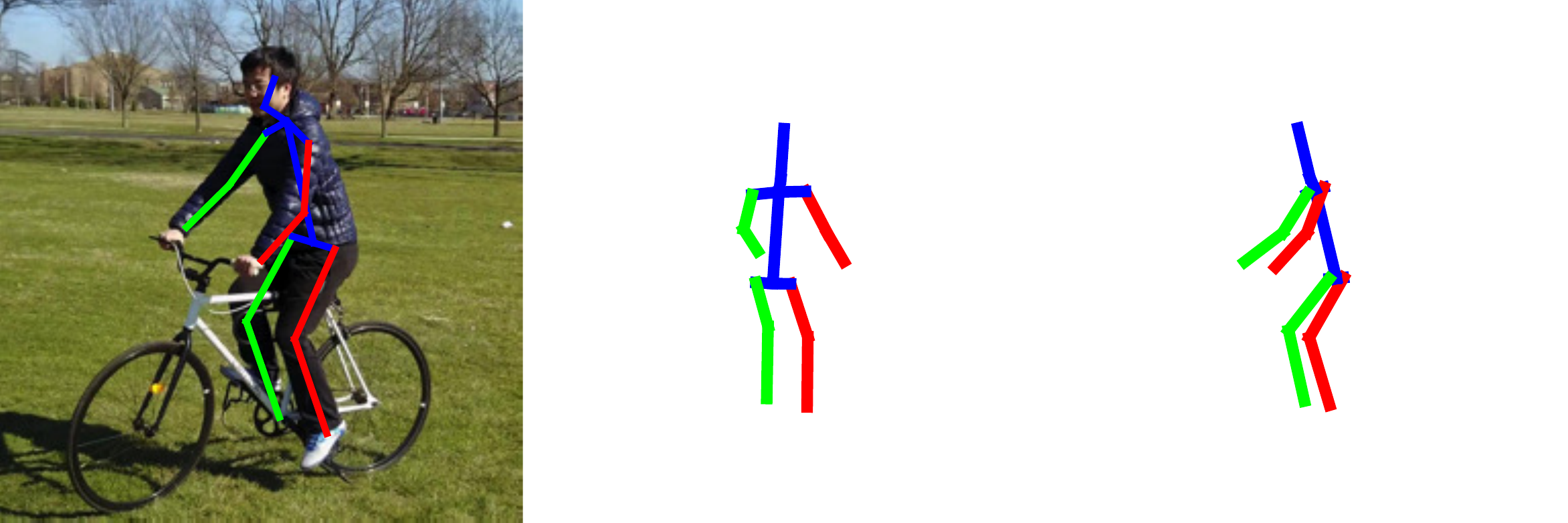}\\
\end{minipage}
\begin{minipage}{0.33\textwidth}
  \includegraphics[width=.99\linewidth]{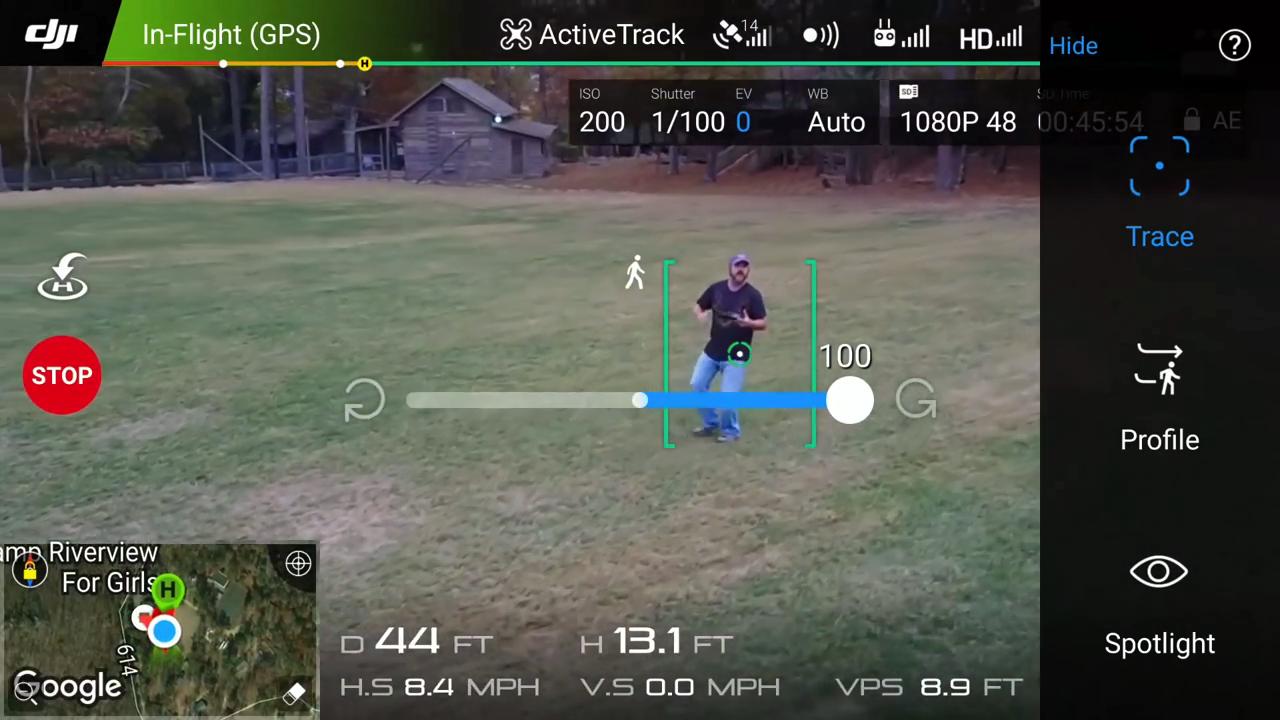}\\
  \includegraphics[width=.99\linewidth]{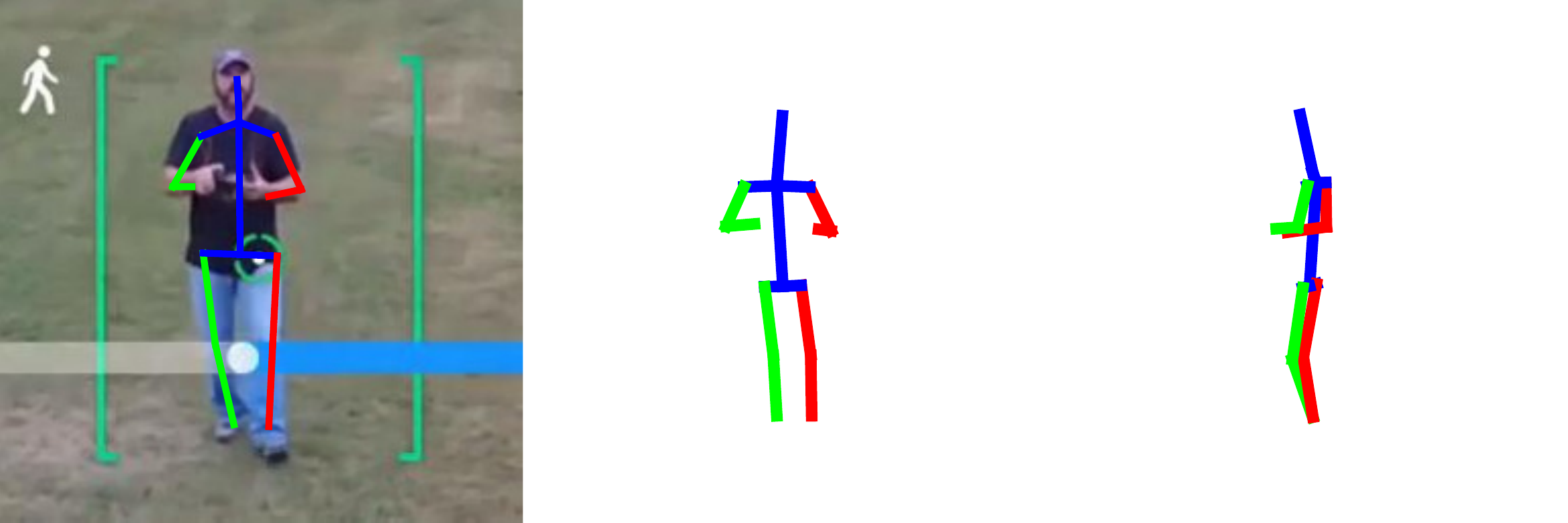}\\
  \includegraphics[width=.99\linewidth]{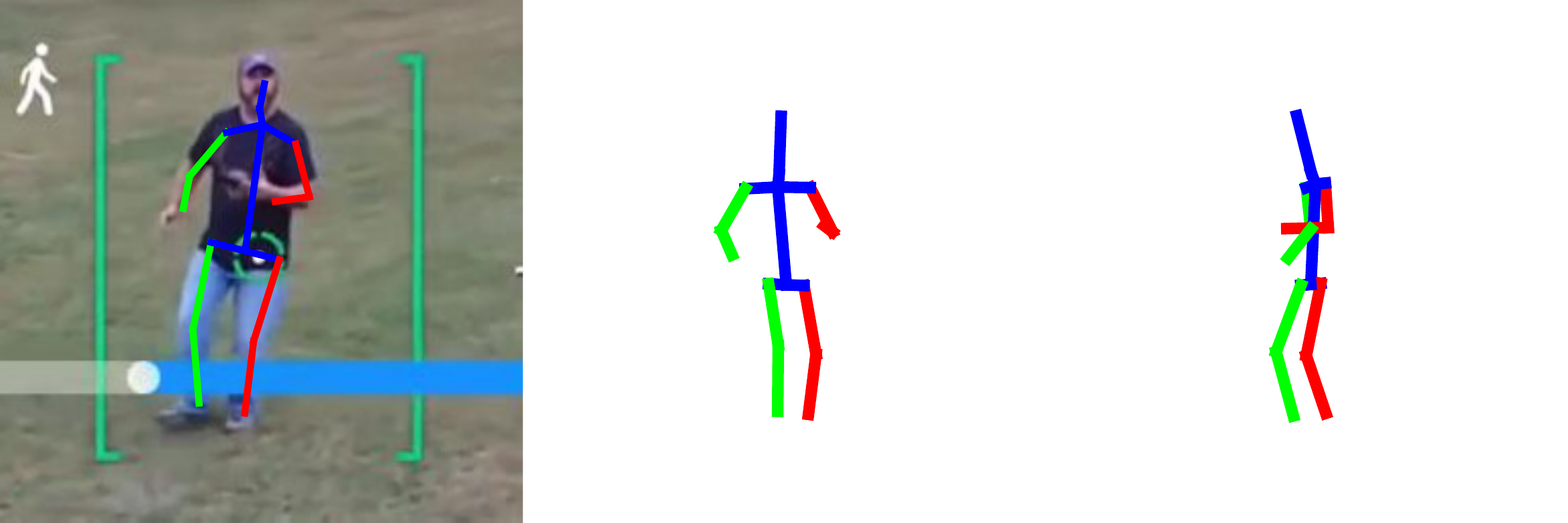}\\
  \includegraphics[width=.99\linewidth]{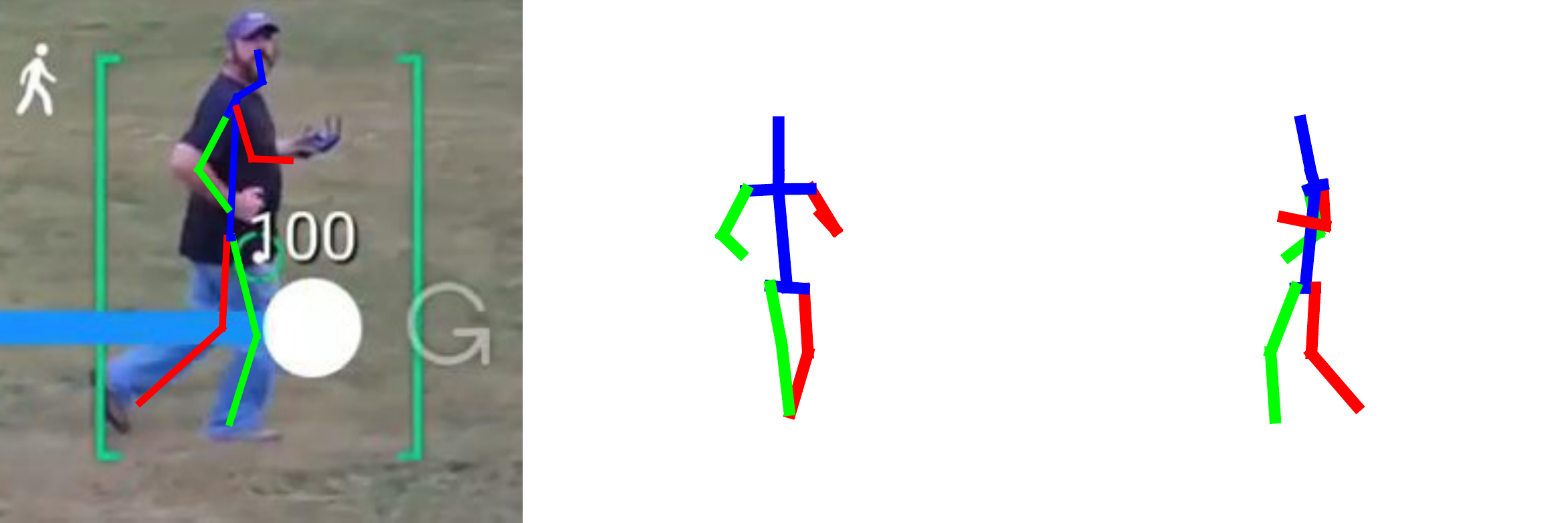}\\
  \includegraphics[width=.99\linewidth]{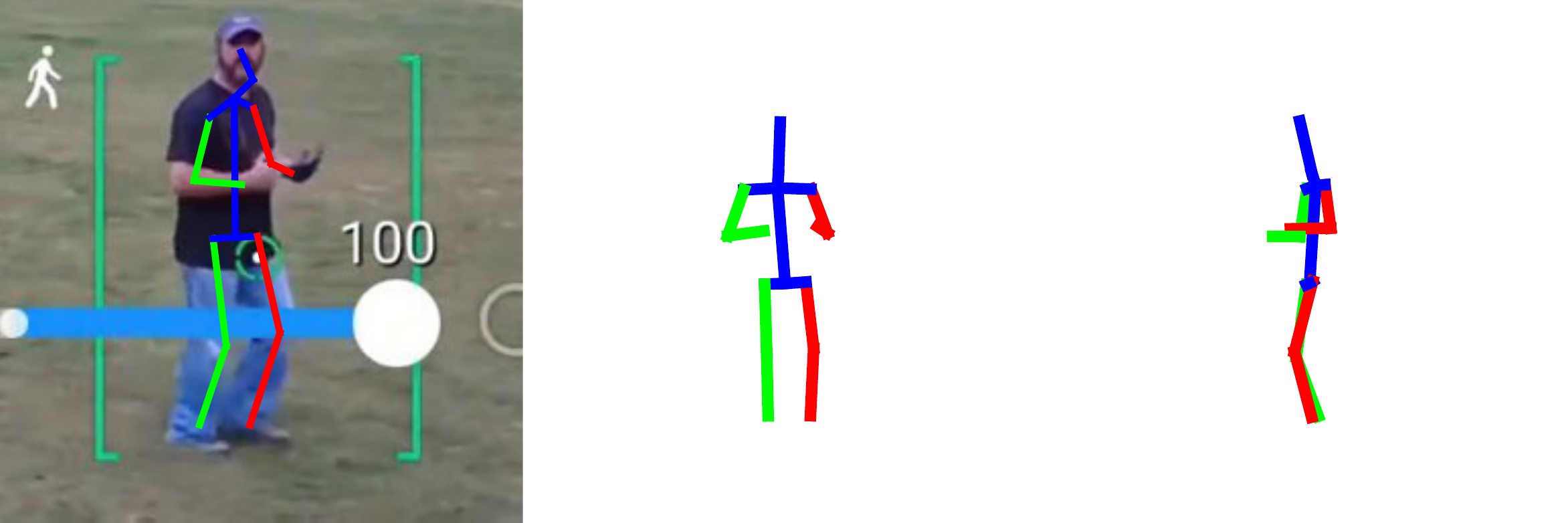}\\
  \includegraphics[width=.99\linewidth]{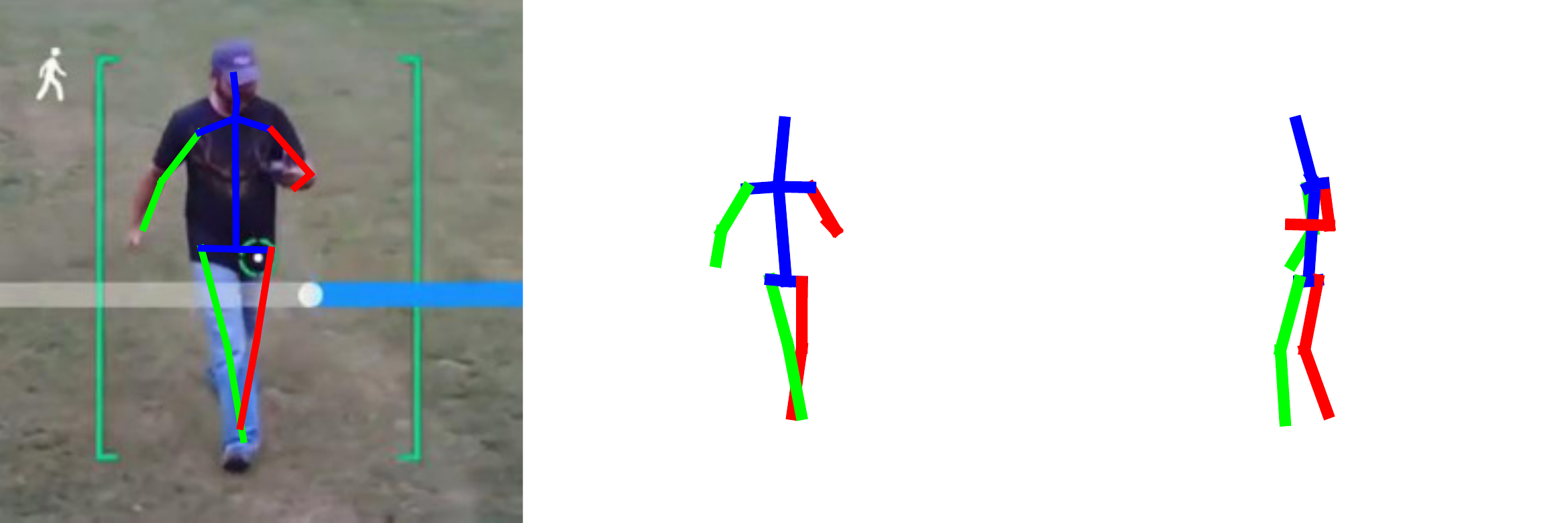}\\
\end{minipage}
  \caption{Reconstruction from outdoor sequences captured by DJI Mavic Pro. An original frame of each sequence is shown in the top row. The following rows correspond to several selected frames, showing the estimated 2D pose in the cropped image and the 3D pose visualized from the front and profile views. The last sequence is from a YouTube video \cite{mavic}. }\label{fig:mavic}
\end{figure*}

\subsection{DroCap dataset}

We collect a new drone-based MoCap dataset named DroCap. The ground truth of body motion is provided by a Vicon MoCap system with markers attached to body joints. We mount a GoPro camera on the AscTec Pelican quadrotor platform (\refFig{fig:lab}) and program it to autonomously track a pre-defined trajectory which is a circle centered at the subject and the desired orientation of the camera is always pointing at the center. The desired speed is 1m/s corresponding to an angular velocity around $25^\circ/s$. During data collection, the subject is doing a variety of actions, such as walking, boxing, and playing soccer, staying at the same location due to the limited indoor space. The current dataset consists of 6 sequences from 2 subjects. 

The 3D human poses reconstructed from the monocular videos are compared to the ground truth obtained from Vicon. Note that no training data is provided. For the proposed approach, the stacked hourglass model \cite{newell2016stacked} trained on MPII \cite{andriluka20142d} is adopted for 2D pose estimation and a pose dictionary learned from Human3.6M \cite{ionescu2014human} is used for single-frame initialization. 

The qualitative results on several representative frames are visualized in \refFig{fig:DroCap}. While the initial single-view estimates by \cite{zhou2016sparseness} have captured global structures, the reconstructions after the multi-frame bundle adjustment are closer to the ground truth recovering more faithful details, e.g., the joint angles of elbows or knees. The bottom-left figure in \refFig{fig:DroCap} shows an example where the original 2D pose estimate is inaccurate but the final reconstruction is correct after handling 2D uncertainties by \cite{zhou2016sparseness}. 

\begin{table*}[t]
\caption{The mean reconstruction errors (mm) on the DroCap dataset.}
\label{tab:DroCap}
\centering
\renewcommand{\arraystretch}{1.2}
\begin{tabular}{lccccccc}
\toprule
& Box1 & Box2 & Walk1 & Walk2 & Soccer1 & Soccer2 & Mean \\
\toprule
MF+NNM \cite{dai2012simple}       & 57.3 & 86.4 & 78.1 & 63.2 & 123.9 & 93.2 & 83.7 \\
Initial \cite{zhou2016sparseness} & 74.0 & 86.7 & 62.0 & 77.0 & 75.7  & 78.4 & 75.6 \\
Initial+BA & \textbf{53.9} & \textbf{70.6} & \textbf{41.1} & \textbf{47.2} & \textbf{56.3}  & \textbf{62.6} & \textbf{55.3} \\
\toprule
\end{tabular}
\vspace{-2em}
\end{table*}

\begin{table}[t]
\caption{The mean reconstruction errors (mm) for different joints.}
\label{tab:joint-error}
\centering
\renewcommand{\arraystretch}{1.2}
\begin{tabular}{*{7}{c}}
\toprule
Wrist & Elbow & Shoulder & Hip & Knee & Ankle \\
\toprule
70.4 & 62.8 & 39.1 & 39.5 & 57.6 & 62.2 \\
\toprule
\end{tabular}
\vspace{-2em}
\end{table}

The reconstruction errors at 12 joints (wrists, elbows, shoulders, hips, knees and ankles) are evaluated. The mean reconstruction errors for each sequence are given in \refTab{tab:DroCap}. ``Initial" and ``BA'' denote single-frame initialization and multi-frame bundle adjustment, respectively. 
A baseline method ``MF + NNM'' is included in comparison, where the 2D joint tracks detected by the same CNN-based detector are input to the state-of-the-art NRSFM method, i.e., matrix factorization for initialization followed by nuclear norm minimization for structure refinement \cite{dai2012simple}. The proposed approach outperforms the baselines, achieving an average error around 55mm. The performance gain over the single-frame initialization is mainly due to the existence of fast camera motion in drone-based videos that provides richer information for reconstruction. 
Moreover, the gain is more significant for more repetitive motions such as walking as the pose sequence can be better represented by a low-dimensional subspace. 
The errors for separate joints are presented in \refTab{tab:joint-error}.  

\subsection{Outdoor MoCap}

Finally, we demonstrate the applicability of the proposed system for outdoor MoCap using consumer drone DJI Mavic Pro. The built-in active tracking function on Mavic Pro is used to track and orbit the moving subject autonomously.
Several example sequences are shown in \refFig{fig:mavic}, including a YouTube video \cite{mavic} to demonstrate the generalizability of the proposed algorithm. The same as previous experiments, no additional training is used. The generic stacked hourglass model \cite{newell2016stacked} trained on MPII \cite{andriluka20142d} is used for 2D pose estimation and the pose dictionary learned from Human3.6M \cite{ionescu2014human} is used for single-frame initialization. The reconstruction results for several selected frames are shown in \refFig{fig:mavic}. 
As shown, the details of the subject motion are well captured. For example, we can clearly see in the last sequence that the right arm of the subject is swinging while the left hand that holds the remote controller is relatively static.

\subsection{Running time}

The reconstruction algorithm was running offline on a desktop with an Intel i7 3.4G CPU, 8G RAM and a GeForce GTX Titan X 6GB GPU.
The running time per frame was $\sim$0.2s for 2D pose estimation and $\sim$0.3s for single-frame initialization, which could be easily paralleled. For a sequence of 300 frames, the running time for multi-frame bundle adjustment was $\sim$8s.

\section{Discussion}

We proposed a novel system for human MoCap using an autonomously flying drone, aimed to address limitations of existing MoCap systems that rely on markers and static cameras. The proposed system is applicably both indoors and outdoors and is capable of using a consumer drone without the need of particular system calibration or model training. We also introduced a new dataset for drone-based MoCap. This work is an initial effort towards drone-based MoCap, which can be potentially extended, e.g. using multiple drones or active trajectory planning, for more accurate reconstruction.

\bibliographystyle{IEEEtran}
\bibliography{bibref_definitions_short,bibref}

\end{document}